# From *n*-grams to trees in Lindenmayer systems

Diego Gabriel Krivochen

diegokrivochen@hotmail.comAbstract:

In this paper we present two approaches to Lindenmayer systems: the rule-based (or 'generative') approach, which focuses on L-systems as Thue rewriting systems and a constraint-based (or 'model-theoretic') approach, in which rules are abandoned in favour of conditions over allowable expressions in the language (Pullum, 2019). We will argue that it is possible, for at least a subset of Lsystems and the languages they generate, to map string admissibility conditions (the 'Three Laws') to local tree admissibility conditions (cf. Rogers, 1997). This is equivalent to defining a model for those languages. We will work out how to construct structure assuming only superficial constraints on expressions, and define a set of constraints that well-formed expressions of specific L-languages must satisfy. We will see that L-systems that other methods distinguish turn out to satisfy the same model.1. Introduction

The most popular approach to structure building in contemporary generative grammar is undoubtedly based on set theory: Merge creates sets of syntactic objects (Chomsky, 1995, 2020; Epstein et al., 2015; Collins, 2017), such that Merge(A, B) results in the set {A, B}; in some versions where A always projects a phrasal label, the result is {A, {A, B}}. However, the primacy of set-theory in generative grammar has not been uncontested: syntactic representations can also be expressed in terms of *graphs* rather than *sets*, with varying results in terms of empirical adequacy and theoretical consistency. Graphs are sets of nodes and edges; more specifically, a graph is a pair G = (V, E), where V is a set of vertices (also called *nodes*) and E is a set of edges; $v \in V$ is a vertex, and $e \in E$ is an edge. An edge *e* joining vertices *a* and *b* is notated *e<a, b>*, and *a* and *b* are said to be *adjacent vertices*; a graph is *directed* iff *e<a, b> ≠ e<b, a>*. Trees are, technically, specific kinds of graphs. A *tree* T is a graph that has no loops (there is no path in T that begins and ends in the same vertex) and is connected (for every two vertices $v_x$, $v_y$ there is a finite path from $v_x$ to $v_y$ or vice-versa). The graph-theoretic approach to syntactic structure can be traced back to Bach (1964), where phrase markers are defined as '*topological structure*[s] *of lines and nodes*'. This perspective allowed for the formalisation of conditions over structural descriptions in graph-theoretical and geometrical terms (e.g., Zwicky & Isard, 1967; McCawley, 1968; Morin & O'Miley, 1969; Kuroda, 1976; Arc Pair Grammar; Johnson & Postal, 1980 and its spiritual successor, Metagraph Grammar; Postal, 2010). McCawley (1968) is often credited with providing a re-interpretation of phrase structure rules (PSR), rewriting rules of the form X → Y, not as mappings from strings to strings (*à la* Chomsky, 1959), but rather as node admissibility conditions (NAC) in graphs. Let us flesh this out. Consider the PSR A → BC. Then,

> *the base component is a set of node admissibility conditions, for example, the condition that a node is admissible if it is labeled A and directly dominates two nodes, the first labeled B and the second labeled C.* (McCawley, 1968: 247)

This view has been taken up by Gazdar (1981); Sag et al. (1985), and others. Pullum (2019) is rather critical of this approach, however:

> *NACs are not conditions on the admissibility of nodes or trees or anything else. An NAC saying 'A → B C' doesn't place any condition on nodes, not even on nodes labeled A: it requires neither that a node labeled A should have the child sequence B C (there could be another NAC*

> *saying A → D E F) and it doesn't require that a child sequence B C must have a parent node A (there could be another NAC saying 'D → B C').* (Pullum, 2019: 62)

We do not agree with this criticism. Suppose that a grammar, conceived of as a string recogniser, is implemented in a formal automaton, such that a rule specifies the state in which the automaton is at a given time and the state to which it proceeds given a certain input. Then, a deterministic pushdown automaton (DPDA) could contain rules like A → B C and A → D E F and the system would still be deterministic, since the interpretation of these rules involves being in state A, getting a specific input, pushing something to the stack, and proceeding to the next state; the input symbol and the symbol pushed to the stack are different in each of those rules. But this is not the way in which PSRs -or, indeed, grammars more generally- are conceived of in McCawley (1968) or Sag et al. (1985). To use Chomsky's (1956) notation, if a phrase structure grammar is deterministic, and contains a rule A → B C, this means that a symbol A in line $\varphi_i$ of a derivation (a string of symbols from the alphabet) can only be replaced by B C in line $\varphi_{i+1}$ (incidentally, no other symbol can be replaced in the transition from line $\varphi_i$ to line $\varphi_{i+1}$). There could be, of course, another rule A → D E F, but in that case we would be introducing non-determinism since a symbol A in line $\varphi_i$ could be replaced by the sequence D E F *or* by the sequence B C in line $\varphi_{i+1}$. For example, Sag et al. (1985: 127) specify that given a rule like A → B C D,

> *This rule specifies part of the conditions that must hold of a structure rooted in A: namely, that it consist of exactly three daughters whose categories are B, C and D, respectively. However, it does not in itself say anything about the linear order in which B, C and D must occur under A.*

The last remark about order will become very relevant below, but for now let us focus on the conditions imposed on A. There is nothing in McCawley's work that leads one to think that he admitted the kind of non-deterministic rules that we have just discussed in his model of the base component of a transformational grammar. That means, at worst, that McCawley's proposal cited above should be rephrased as follows:

> *…the condition that a node is admissible **in a deterministic path on a tree T described by a deterministic CF grammar iff**…*

In which case, given a rule A → B C, a node A in a tree T can *only* dominate B and C (as in Sag et al.). Of course, this does not put any condition on B and C themselves. Consider the following PSRs:

1) S → NP VP
   VP → V (NP)

What this is saying (in the McCawlean interpretation) is that a node S in a tree T is admissible *iff* it immediately dominates nodes labelled NP and VP. But it does *not* say that a node NP is only admissible in T if immediately dominated by S; as a matter of fact, the grammar in (1) perfectly allows for a node labelled NP to be dominated by a node labelled VP as well as by a node labelled S. Furthermore, and against early (1955) Generative practice (but in tune with the model in Chomsky, 1965), phrase structure rules of the kind A → φ A ψ are allowed (i.e., recursion is a property of the base component, not of the transformational component). This means, for our present intents and purposes, that a node *n* labelled A may transitively dominate another node *m* also labelled A in a well-formed tree T, for *n* ≠ *m* (note: the *nodes* are -or may be- distinct, their *labels* are not). While we have nothing against loops in graphs (they can be used to great effect in grammatical description), in this particular case we will not allow for a node in T to directly dominate itself in any path defined in T; a node may, however, immediately dominate another node which is assigned to the same indexed category (see Postal, 2010 for some discussion about this point).

There seem to be some arguments in favour of keeping the idea that PSR may be interpreted as NACs, at least for grammars in canonical form. However, to our knowledge, this approach has not been applied to other kinds of formal systems. In this work, we will look at how a view of a grammar as a set of NACs can assist us in the analysis of other types of formal systems, in particular, a special kind of Lindenmayer system that has been used fruitfully in linguistic and psycholinguistic research (Saddy, 2009; Phillips, 2017; Vender et al., 2020). The questions we aim to answer are (a) whether we can extend the view of a grammar as a set of NACs from grammars in canonical form to Lindenmayer-systems, and (b) whether there is any way to link superficial properties of strings to NACs in trees (i.e., if we can use this framework to have a mapping from strings to graphs).

2. <u>NACs in (some) Lindenmayer systems</u>

Lindenmayer systems (L-systems henceforth; see e.g., Lindenmayer, 1968; Prusinkiewicz & Lindenmayer, 2010; Rozenberg & Salomaa, 1980) are recursive rewrite systems characterised by some of the same components that define systems in Chomsky-normal form: an alphabet and a set of rules over the alphabet. L-systems, however, differ from rewrite systems in Chomsky-normal form in two fundamental respects:

a. There is no distinction between terminal nodes and nonterminal nodes in the alphabet of the grammar[1]
b. There is no sequentiality in rule application: all rules that can apply do so simultaneously

It is important to separate, when looking at a grammar, properties of the *outputs* of that grammar ('representational properties') from properties of the relations between outputs ('derivational properties); in other words, *states* versus *processes*. Derivationally, it must be noted that the simultaneity of rule application in L-systems contrasts drastically with the sequentiality of rule application in normal grammars. While only one rule can apply at a time *per generation* in a Chomsky-normal grammar, even if there is more than one nonterminal that can be rewritten[2], L-grammars rewrite *all* possible symbols *per* generation with all rules that can apply doing so *at the same time*, yielding a completely different growth pattern. We can exemplify the two distinctive properties of L-systems in the derivation below, corresponding to the so-called XOR grammar (Saddy, 2009; Shirley, 2014):

2) Alphabet: *a*, *b*
   Rules: *a* → *a b*
          *b* → *b a*
   Axiom: *a*
   Derivation:

   $$a$$
   $$ab$$

---

[1] The alphabet of a grammar in Chomsky-normal form contains two sets of symbols, call them $V_N$ and $V_T$, which constitute the non-terminal and terminal vocabulary of the grammar respectively. Hopcroft & Ullman (1969: 10) explicitly say that 'We assume that $V_N$ and $V_t$ contain no elements in common'. Similarly, Levelt (2008: 4) says '$V_N$ and $V_T$ are disjoint: their intersection, $V_N \cap V_T$, is empty'. In classical L-systems (Lindenmayer, 1968; Prusinkiewicz & Lindenmayer, 2010), every symbol may appear at the left-hand side and at the right-hand side of a rule, thus effectively dissolving the $V_N$, $V_T$ distinction.
[2] In this respect, it is useful to refer to Lees' (1976) analysis of the formal conditions over *immediate constituent* approaches to structural descriptions, which are the basis for generative grammars, both transformational and non-transformational (see Schmerling, 1983 for further discussion about *immediate constituent* grammars). He concludes that the essential condition for the formulation of the rules of a grammar '*is simply that no more than one abstract grammatical symbol of a string be expanded by a given rule at a time*' (Lees, 1976: 30)

*abba*
*abbabaab*
*abbabaabbaababba*
...

In principle, there is nothing in L-systems that prevent us from conceiving of the rules of an L-grammar as NACs: this will be crucial in our proposal to transition between superficial regularities and local derivational objects. Consider, for instance, the rules of the so-called Fibonacci grammar[3], and how they should be read if interpreted as NACs:

3) $0 \rightarrow 1$ (a tree T with a node labelled 0 is well formed iff every node labelled 0 immediately dominates a node labelled 1 in T)
$1 \rightarrow 0\ 1$ (a tree T with a node labelled 1 is well formed iff every node labelled 1 immediately dominates a node labelled 0 and a node labelled 1 in T)

Crucially for our purposes, the biconditional follows the relation of *dominance*: as above, where an NP could appear in two configurations (dominated by S and dominated by VP -and presumably in a variety of other configurations, like dominated by PP-), here a node labelled 1 may appear in a tree T dominated by either 1 or 0; but it can only dominate a 0 *and* a 1[4]. Krivochen et al. (2018) identify some important differences at the level of constituency between the Fibonacci (Fib henceforth) grammar and the grammar that results of inverting the linear order of the terms dominated by 1 (i.e., the second rule becomes $1 \rightarrow 1\ 0$), which is dubbed *bif*. Regardless of their differences at the level of constituency and the possibilities of reconstructing structure on the basis of local relations, superficially the strings generated by both grammars are remarkably similar. A Fib and a bif derivation look as follows:

4)  
| Fib: | | bif: | |
|---|---|---|---|
| | 0 | | 0 |
| $0 \rightarrow 1$ | 1 | $0 \rightarrow 1$ | 1 |
| $1 \rightarrow 0\ 1$ | 01 | $1 \rightarrow 1\ 0$ | 10 |
| | 101 | | 101 |
| | 01101 | | 10110 |
| | 10101101 | | 10110101 |
| | 0110110101101 | | 1011010110110 |
| | … | | … |

---

[3] The Fibonacci grammar (so called because the number of total symbols at each line of the derivation, as well as the number of occurrences of each symbol is always a Fibonacci number) has been used in a variety of linguistic contexts, most notably in connection to a specific interpretation of X-bar theory (Uriagereka, 1998; Medeiros & Piattelli-Palmarini, 2018; among others). The present discussion bears no relation to that literature. Even though it is possible that the study of L-systems may inform some aspects of natural language syntax (and even then, the universal applicability of endocentric, binary-branching, single-rooted phrase markers has been empirically challenged), the present work is concerned with the relation between the study of superficial regularities and the development of adequate models in a kind of system that does not feature some prominent formal properties of natural language grammars (most notably, the distinction between terminal and non-terminal nodes). An extension of the present system to natural languages would require additional restrictions over allowable trees and the conversion of an L-system into a grammar in canonical form.

[4] The conjunction of the NACs (McCawley, 1968: 248) is locally satisfied by so-called *k*-points in the Fibonacci grammar: these are 1s which are dominated by 0 and which dominate 0 1. *K*-points are defined in Krivochen et al. (2018) as labels for local units, since they provide information about their neighbourhood: identifying a point as a *k*-point entails identifying the points in its neighbourhood (the 0 that dominates it, the 1 and 0 that it dominates). An inclusive disjunction of NACs includes *n*-points (1s dominated by *k*-points, thus part of the neighbourhood of *k*-points) and *s*-points (1s dominated by *n*-points, thus outside the neighbourhood of *k*-points).

The Fib grammar delivers certain superficial regularities, in terms of precedence relations in the strings that are produced by applying the rules; in the case of Fib we have referred to these regularities as the Three Laws (Krivochen et al., 2018):

5) First Law: every 0 is followed by a 1 (*00)
   Second Law: two 1s are always followed by a o (*111)
   Third Law: a single 1 may be followed by either a 0 or a 1

Here we will focus on the First and Second Laws (the two deterministic Laws), which are conditions over 2- and 3-grams for the Fib grammar and its sibling grammar *bif*. In previous works we have explored the structure of these grammars, and the way in which local structure can be reconstructed based on specific designated nodes in the tree (so-called *k*-points in Krivochen et al., 2018, Vender et al., 2020); the present paper has a different focus: we will present a method that, given the superficial regularities that arise in the Fib and *bif* grammars (the First and Second Laws), allows us to approximate possible structural descriptions for strings.

The main idea is the following: we can think of the First and Second Laws as NACs for trees of depth 1. That means that for the First and Second laws we can define the complement set of strings allowed, for 2- and 3-grams (since these are the *n*-gram sizes that the Laws refer to):

6) {11, 01, 10}
   {101, 110, 011, 010} {*00} is excluded by the 1$^{st}$ Law and therefore all 3-grams containing it also will.

Recursively, this applies to any string that properly contains any of these *n*-grams (thus, for instance, if the tri-gram *111 is not allowed, then any *n*-gram containing *111 will be also not allowed, for all values of *n*).

What we have now is a way to obtain a description of trees over an alphabet Σ = {0, 1} which follow specific NACs. Let $T_b^d$ be a tree with depth *d* and breadth *b*. Following Pullum (2019: 64), we can define NACs of depth zero for each terminal symbol in the alphabet. Because there is no difference between terminals and non-terminals in L-systems, each symbol in the alphabet can be both the root and the frontier of a tree. Then, the complement of the set of strings forbidden by the First Law should define exhaustively the set of trees with breadth 2 and depth 0 over Σ. And the complement set of the Second Law does the same for trees with breadth 3 and depth 0 over Σ. Since we are only dealing with depth 0 for the time being we will just use the subindex for breadth (thus, the First and Second Laws define the sets $T_2$ and $T_3$). Call the trees defined by the interpretation of the Fib rules as NACs, *elementary trees* (see Joshi, 1985 for a related use of this term).

Then, we have a set of expressions (i.e., strings). What we want is to be able to use superficial regularities (properties of expressions) as a way to obtain a partial characterisation of sets of trees which make reference to nothing else than elements in the string. In order to do this we can operate only with the restrictions defined above. We are able to consider only restrictions over expressions (as opposed to formulate rules that produce or enumerate well-formed expressions) because, as Pullum (2019) points out, making sure that a set of trees does *not* generate (in the sense of 'produce as terminal strings') the 2-gram *00 and 3-gram *111 is equivalent to building a constructivist theory in which all the other 2- and 3-grams over the same alphabet are indeed allowed. But strings are longer than 2- and 3-grams: we need a way to combine basic expressions to obtain derived expressions which belong to indexed categories of the grammar, which means, presumably, that we need bigger trees than elementary trees as well. Checking whether an expression belongs to a category of the grammar has the form of a set of *if…then* statements (Montague, 1974), and if we can use the same format for NACs then checking expressions and checking local trees follows the same kind of procedure (this is one of the attractive features of model-theoretic syntax). Just like we need to allow for derived expressions in order to satisfy the most basic requirement of observational adequacy, we will allow for elementary trees to be composed, by means of *substitution* (at the root and at the frontier). This implies that the system must distinguish between the indexed categories of the language and identify when two expressions share the same category. In the present context, this last condition implies that the system can establish identity between nodes with the same label. Straightforwardly: if we have two category indexes, 0 and 1, then any 1 (any tree of any depth with root 1) may substitute for any other 1, and any 0 (any tree of any depth with root 0) may substitute for any other 0. In Krivochen (2018) we referred to this property as *perfect structure preservation*. This is a property that L-systems display, due to their lack of distinction between terminal and non-terminal nodes, as opposed to natural languages, in which *structure preservation* needs to make reference to specific configurations (see Emonds, 1970).

The main objective of this work is to create a kind of model-theoretic grammar for Fib, based on constraints over expressions *as well as* tree composition. But tree-composition here will be defined indirectly, through *string concatenation*: every symbol in a string can be trivially taken as the root of a tree of depth 0; less trivial results emerge once we consider co-occurrence restrictions over $n$-grams. We will use these restrictions as a way to build bigger (but not too big) elementary trees. The reliance on superficial co-occurrence restrictions means that we are taking away the inherently derivational character of L-systems, but since we are interested in a model of elements and relations and a set of filters over expressions (and not in procedural proof-theoretic derivations), this is not a problem. The advantages of doing this is that it would allow for an integration between $n$-gram research and structure-based research about how information about structure is extracted from signals or strings (see, e.g., Saddy, 2018) in a way that does not rely on quantitative measures or statistical analysis.

Let us make the proposal explicit. The Second Law allows for the 3-gram

7) 101

And the First Law allows for both 2-grams

8) a. 10

and

b. 01

These are the two 2-grams that we find in [101]. Now, we can put this string in a context: say, preceded by [11], which is one of the 2-grams allowed by the First Law. The result is

9) *11101

which is ungrammatical. However, we do not need to stipulate a constraint over 5-grams to capture this, only considering the First and Second Laws is enough, as noted above.

This description can be made more explicit and powerful still, if we allow for Boolean connectives between string admissibility conditions (that is, conditions over expressions) just like they have been conceived of for NACs (for the latter, see McCawley, 1968). It is easy to see that the sets of allowed strings are *not* closed under concatenation (Boolean AND)[5]:

---

[5] This set is also not closed under Kleene star, but this will not be relevant in the present context.

| *01-11   | *10-010  | *010-011 |
| *10-01   | *011-11  | *010-010 |
| *11-10   | *011-10  | *011-110 |
| *11-11   | *101-11  | *011-101 |
| *01-110  | *110-010 |          |
| *10-011  | *110-011 |          |

If concatenation of *n*-grams is interpreted as Boolean AND, then for the result of the concatenation to be a well-formed expression of the language both *n*-grams must be well-formed expressions of the language themselves.

Concatenation is an interesting relation, in particular in the present context. Consider the consequences of adding a binary predicate *p* that takes *n*-grams *s* and *s'* as arguments and outputs {*s*, *s'*}. How are *s* and *s'* linearly organised? In principle, there is no formal reason to prefer the relation *precedes* to the relation *follows*. And this is important, because the First Law is based on this: 0 cannot follow or precede 0. The Second Law can also be expressed in these terms, considering that we allowed for Boolean connectives: a 1 cannot follow or precede a 1 that follows or precedes a 1. The language $L^2_{K,P}$ used in Rogers (1997)[6], or any such formalism, would allow us to put this in the following terms (more or less)[7]:

10) Let *x* be an indexed category of the language, and let ≺ be the binary predicate *precedes*. Then,
$\neg \exists (x_1) [(x_1 \prec x_2) \rightarrow (x_2 \prec x_3)]$

This is important, because in addition to being a set of conditions on strings, it is also a set of conditions on tree composition. Just like we can decompose a string in *n*-grams, we can decompose trees into treelets, whose minimal depth and breadth are defined by the conditions on string acceptability above: minimal depth 1 (for it to be a non-trivial tree), minimal breadth 2 (since the smallest *n*-gram that the Laws make reference to is a 2-gram). What is the relevance of these conditions over substrings for trees? Remember that we are dealing with trees of depth 1, and that

> *If a node n has at least one descendant other than itself, and has label A, then **A must be in $V_N$**.*
> (Hopcroft & Ullman, 1969: 19)

---

[6] This language is a *second order* language (thus, there is quantification over predicates), with a set of individual constants K, a set of one place predicates P (labels), and a set of two-place predicates (dominates, properly dominates, precedes, is-equal-to).

[7] It is important to note that the theory used, for instance, in Montague (1974) [1970] has pretty much the same form as Rogers'; no rewriting rules or phrase structure trees at all, only *if…then* conditions over expressions (we mention EFL and not PTQ because the rule of *quantifying in* present in the latter work pushes the power of the system beyond strict context-free, and if we want to make use of Rogers' apparatus, then we better stay CF). Proof-theoretic applications of phrase structure grammars very rarely if ever (with the possible exception of GPSG and HPSG, both of which are *model-theoretic* rather than *proof-theoretic* in Pullum's sense) reached the level of explicitness that Montague grammar did. The point is, building a grammar from expressions up is thus not only possible but also perhaps desirable (see also Schmerling, 2018 for a recent development in this line).

This means that 1 and 0 are in $V_N$ as well as in $V_T$, since there are no constraints on labels for nodes in trees implicit or explicit in the Laws. Given the fact that we have only two indexed categories and all rules make reference to both, there is no object that excludes a category.

Consider now the tree (11a), constructed using sub-trees (*elementary trees*) (11b) and (11c):

11) 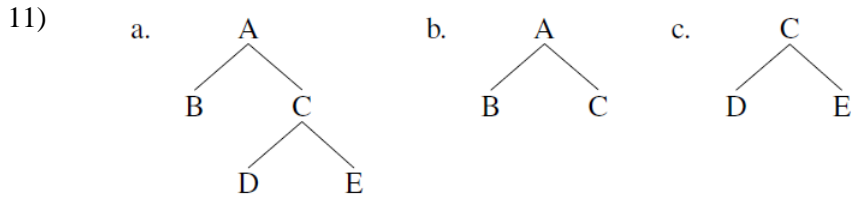

There is an overlap of a symbol: the root of a tree (node C, in (11c)) is the frontier of another (the same node in (11b)); this is simply a case of *substitution* à la Chomsky (1955). It is important to note that in principle (and particularly in the context of L-systems) there is no *a priori* condition against substitution *at the root* rather than *at the frontier*: if there is a node R in trees $T_1$ and $T_2$, such that R immediately dominates A, B, C in $T_1$ and D, E, F in $T_2$, then we can create tree $T_3$, in which R immediately dominates A, B, C, D, E, F.

So far we have been dealing mainly with conditions over (sub-)*strings*; we can now introduce some conditions on *trees* (see e.g. Rogers, 1998: 17-18 for a formal specification of tree axioms for CFGs):

I. Every node (other than the root) has a mother
II. Every mother has at least one daughter
III. If a node has *m* daughters in a treelet T and *n* daughters in a treelet T', for $n < m$, it will have exactly *m* daughters in every T" that properly contains T or properly contains T' (*maximise connectivity, minimise number of nodes*)

The first and second conditions, in the present context, require us to ask the questions: 'what symbols from the alphabet can be the first argument of the relation *mother-of*?' and 'what symbols from the alphabet can be the second argument of the relation *mother-of*?'. In grammars in canonical form, the nodes in the set $V_N$ can always be the first argument of the relation *mother-of*, and the nodes in the sets $V_N$ (minus the designated root symbol) and $V_T$ can be the second argument. L-systems, however, do not distinguish between these two sets, which means that the sets of symbols that can be the first and second arguments of the binary relation *mother-of* (a.k.a. *dominates*) are co-extensive: {0, 1}. Unless, that is, there is an independent constraint that we need to consider. We will come back to this shortly. The third condition must be independently justified: what we are saying is that if a node may occur as the root of *n* trees of depth 1, given a set of trees $S = \{T_1, T_2, T_3, …T_n\}$, then the composition of trees from S will always prefer the elementary trees with *greatest* breadth. The condition that these must obey, however, is that the resulting strings do not violate the SACs: recall that the objective is to characterise a string language in terms of a tree set and vice-versa (or at least get a reliable mapping between them). Suppose that we wanted to make the binary relations *precedes* (in a string) and *dominates* (in a tree) isomorphic. Exactly how, is not clear. It may be stipulated, of course (as in certain forms of so-called *antisymmetry theory*; see Kayne, 1994 and much related work), but that would not provide us with much insight since we would be stating a relation that should be derived in this context. One possibility is to define a *walk* for each tree. A $v_1$-$v_2$ *walk* in a directed tree T is a finite ordered alternating sequence of vertices and edges that begins in $v_1$ and ends in $v_2$. Then, if A (immediately) *precedes* B in a *walk* in T, then we can say that A (immediately) *dominates* B in T. We have just dispensed with *dominance* as a primitive[8] (cf. McCawley, 1968; Sag et al., 1985, both of which define the relations *precedes* and *dominates* independently and as equally basic), which is highly desirable: after all, we are looking for a way to connect *n*-grams to graphs such that conditions over *n*-grams can serve as models for local graphs, and *precedence* (as opposed to *dominance* or other structural relations, like *c-command*) is a relation that can be defined in both kinds of objects.

We still have the problem of adequately characterising Fib trees, since the string-tree conversion procedure whereby the relations *node precedes* and *node dominates* are isomorphic can only generate monotonically growing trees in which each node has *two* daughters: *precedes* and *dominates* are necessarily *two-place* relations. However, in the Fib grammar we have a rule in which a node dominates only one node (0 → 1) and another rule in which a node dominates two nodes (1 → 0 1); this is at the core of the classification of the Fib grammar as *asymmetric* (Krivochen & Saddy, 2016). Let us assume, at this point, that an L-grammar may include something like Pullum's (2019: 69) *Lonely Beta condition*:

LONELY BETA $\equiv_{def}$ $(\exists x)[\beta(x) \wedge (\forall y)[(\beta(y) \Rightarrow (y = x)) \wedge (\neg\beta(y) \Rightarrow \alpha(y))]]$
*'There is an x that is labeled β, and x is the only node labeled β (i.e., any y labeled β is identical with x), and any other node (i.e., any y not labeled β) is labeled α.'*

Of course, intuitively we know that our lonely beta is the label '0', assigned to nodes in the tree. This condition can also be somewhat formalised as follows. Let *x* and *y* be variables over indexed categories in an algebra (so, not expressions themselves, but indexed categories assigned to expressions; in tree terms, not nodes, but labels of nodes). Define a binary relation ρ(*x*, *y*) (there are binary relations allowed in $L^2_{K,P}$, so there is nothing affecting the formal power of the system in terms of the languages it can characterise). Then, we have that:

12) $(x, y) \in \rho$

$(y, x) \in \rho$

$(x, x) \notin \rho$

$(y, y) \in \rho$

We have thus characterised our lonely beta (LB) in a different way: a LB is the only indexed category assigned to nodes in a graph that does not allow for a loop arc (in the sense of Postal, 2010). This is the 'independent constraint' that we referred to above: a node labelled 0 cannot be the first *and* second argument of the relation *dominates*. All other configurations involving the LB are permitted, as long as they do not violate any of the constraints independently derived from superficial regularities; these constraints constitute part of the model that expressions of the language must satisfy. This means that we allow for the following trees of depth 1 and breadth 1, and using [0] and [1] instead of *x* and *y*:

13) Set $T^1_1$:

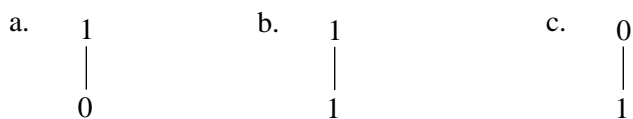

a.  1
    |
    0

b.  1
    |
    1

c.  0
    |
    1

---

[8] This is not to say, of course, that *dominance* has been eliminated altogether; it just is not a primitive notion as it is in X-bar theory and related formalisms. Rather, it follows from the definition of a directed graph.

With the tree

d.   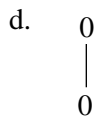

being excluded as an elementary tree: (0, 0) does not belong to the set defined by the relation ρ. But remember that trees can be composed by node identification (root-root or root-frontier): labels are only indexes that allow for this identification. If a node A in T is assigned to the indexed category C, and a node B in T' is assigned to the same category C, when a tree T" is constructed from T and T', A and B can be collapsed as a single node since they are identical for all purposes of the grammar (see also Sarkar & Joshi, 1997). If, again, we allow for free-substitution (such that operations on trees are analogous to graph union with directed graphs) we can construct the following set of trees $T_2^1$ by substitution at the root (we will use ∪ to represent tree composition; note that the order of input trees does matter since we are dealing with directed graphs):

14) Set $T_2^1$

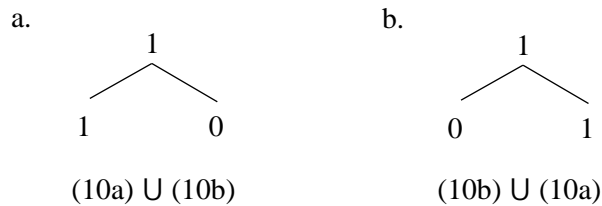

This can be generalised: substitution can target any tree from any set and operate at the root or at the frontier. Therefore, the following trees are legitimate:

15) 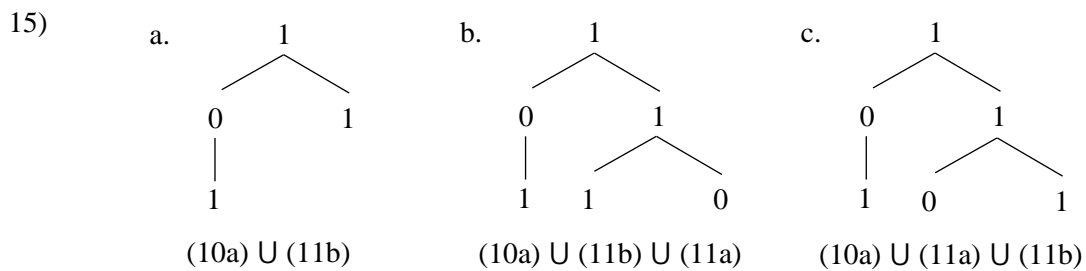

Now, obviously we can in principle have derived trees that look like this:

16) 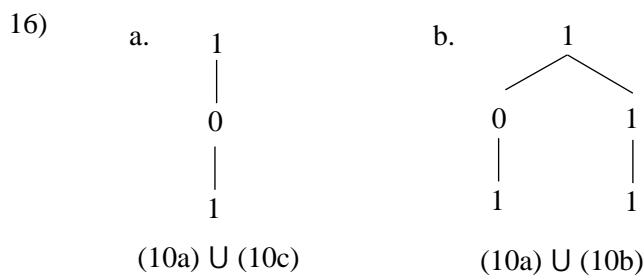

But recall that we have stipulated that

> If a node has *m* daughters in T and *n* daughters in T', for *n* < *m*, it will have exactly *m* daughters in every T" that properly contains T or properly contains T'

This means that given an option for [1] to be the root of $T_1^1$ and of $T_2^1$ as the input for (root or frontier) substitution, the latter (the wider tree) will be preferred in the composition of $T_n^2$, unless the resulting expression violates a string admissibility condition (here, the First or Second laws). At this point, then, we have the trees in (13a) and (14a) and (14b). The language described by these trees is the union of what in other works we have referred to as Fib and bif, depending on whether (14a) or (14b) is chosen consistently (see (15b) and (15c)). The point of this stipulation is that, once a legitimate context for a symbol has been identified, the system aims at maximising the size of that context: an advantage of this is that the same storage space (one elementary tree) can now describe more structure in terms of number of nodes if an elementary tree with root *x* that belongs to $T_1^1$ can be replaced by an elementary tree with the same root that belongs to $T_2^1$. At the same time, elementary trees do not need to be too big (in particular, too deep), since the grammar also contains an operation of tree composition: trees of depth 1 suffice, in the case under consideration here, to define the set of elementary trees; deeper trees can be obtained via substitution. Note that, in this context, all trees with root 1 will dominate at most two symbols, never more; all trees with root 0 will dominate only one symbol as per the LB condition.

3. Space-filling grammars

In this section we will explore a slightly different way to get to the set of trees that satisfy the model defined by the First and Second Law. Restrictions over strings, that is, over 1-dimensional objects, may be extended to 2-dimensional objects if we consider the derivation of a grammar to be a procedure for the parametrisation of a 2-D space (Krivochen, 2018; Saddy, 2018). We may proceed in the Euclidean way: a point (an element in the alphabet) has no dimension, a line (a string) has one, a plane has two. What is the 'plane' here?

Let Λ be a lattice, where each point may be in one of two states. Let those two states be 0 and 1, the alphabet of the formal system we will use to provide a specification of that lattice. Then, we can use the First and Second Laws as a way to specify the 'spin' of each point: 1 or 0 in a 1-D scenario, analogously to how the Game of Life (Gardner, 1970) can be similarly constructed. Mitchell et al. (1993) summarise environmental conditions for the development of 1-D cellular automata: given a binary alphabet Σ = {0, 1}, let *η* be the set of possible neighbourhoods, let *ϕ* be a function that determines the behaviour of a target cell depending on the neighbourhood, and let $s = \phi(\eta)$ be the 'output bits', to which the central cell (bolded) is updated. In the diagram of this GoL, let *time* flow vertically, such that we get strings at each application of *s*. Then, the 1-D GoL rules look like this:

| *η* | 0**0**0 | 0**0**1 | 0**1**0 | 0**1**1 | 1**0**0 | 1**0**1 | 1**1**0 | 1**1**1 |
|---|---|---|---|---|---|---|---|---|
| *s* | 0 | 0 | 0 | 1 | 0 | 1 | 1 | 1 |

Note that the number of symbols per string does not change, we are merely updating the index that each cell is assigned depending on its neighbourhood. In terms of strings, the function *s* is telling us that if [000] is a substring in a generation $g_n$ of the 1-D GoL, then [010] is not a legitimate substring that corresponds to the same place in the string at $g_{n+1}$. In other words, that the central 0 in 000 cannot precede a 1 in the *y* axis (after a single application of the transition function *s*). What we can do now is adapt this line of reasoning to our needs. Remember the possible 2-grams in the language that we are characterising, defined by asymmetric L-systems Fib and bif:

17) {11, 01, 10}

When we consider the First and Second laws as conditions over sub-strings, we can do that in either the *x* or *y* axis, provided we do that one axis at a time. So, 0 cannot precede 0 in either *x* or *y* axis, because we get strings in both directions since everything is a terminal (here L-systems differ from Chomsky Normal grammars, of course). The latter case (*0 ≺ 0 in the *y* axis) amounts to the prohibition of the tree (10d). In this sense, the relation *dominates* (⊳ in Rogers, 1997) can be rephrased as *precedes in the y axis*. We may effectively eliminate *dominance* as a primitive, given our focus on strings. Is it as terrible a reinterpretation as it seems? Not really: after all, we can define a walk through a path defined by a number of nodes and edges: a walk in a graph (Van Steen, 2010). As we said above, if T is a rooted, directed graph, and if A *dominates* B in T, then A will be walked on *before* B in a walk defined for T (this walk may be a trail or a path, depending on whether re-visiting a node is allowed or not). So, nothing has been lost at the level of description we are working with here. The inclusion of the Lonely Beta condition prevents overgeneration: imagine we assumed simply that if A precedes B then A dominates B. Then, every bi-gram should map to a binary-branching tree, Merge-style. But then we would be missing the whole point of asymmetric L-grammars, and mischaracterising the languages they generate. There is a very important point to make here: *all we have, all we assume, is the First and Second Laws*. We have started constructing trees from single symbols, and as a matter of fact, the Lonely Beta condition is not strictly required to capture the intuition that the First Law applies in both axes of the plane parametrised by Fib. It is just a matter of formal convenience (and because a reference never goes amiss).

Importantly, nothing prevents, in principle, that the 'illegal' *n*-grams are obtained via *tree composition*; all we have established is that, if those conditions over *n*-grams are interpreted as NACs, then an *elementary tree* cannot feature them. We have said nothing about constraints on *derived trees* (i.e., trees which are the result of composition). Thus, the Second Law would ban an *elementary tree* like (18):

18) 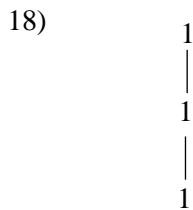

But any *derived* tree which contains (18) as a proper subpart should indeed be permitted. For example, we predict that

19) 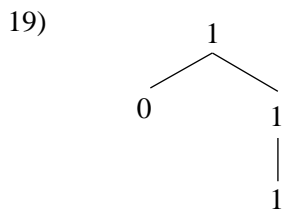

is a permissible local structure under the Second Law. However, that does not mean that (19) is a well-formed tree under current assumptions. The restriction that bans (19) specifically is of a different kind; namely, condition III above. We would then prefer (19'), which respects *all* conditions (the relation specified in (18) is indicated in **bold**):

19') 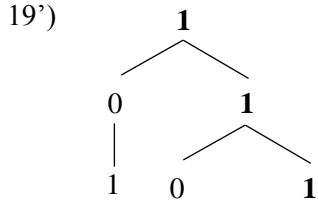

It is also crucial to bear in mind that the fact that (19) is a permissible (but sub-optimal) local structure does not mean that it is a *constituent* of the grammar (a structural unit as defined by the rules of the grammar), that is to be determined by a different procedure. If the rules $0 \rightarrow 1$ and $1 \rightarrow 0\ 1$ are developed in a tree fashion, then (19') actually corresponds to a local description of a constituent, but not (19). If the model that trees need to satisfy is the First and Second Laws, then (19) should be allowed as a local description of possible relation between nodes in a Fib tree; however, it is neither an elementary nor a derived tree.

4. Further restrictions on constituency

The 'different procedure' we mentioned above needs to be specified. In the present context, we can only appeal to the segmentation of the string in *n*-grams since we are trying to get to a set of local structures from a set of (conditions over permissible) *n*-grams: the appeal of *n*-gram analysis is that the size of constituent units need not be known in advance, since we can just shift the size of the target unit from 1 to the length $l$ of the string[9]. There are ways to optimise the *n*-gram selection, as explained in Matlach et al. (2020), but they fall outside the scope of the present work.

The 'alternative' procedure we have in mind (with respect to **Section 2**) goes along the following lines. Take any Fib generation, e.g.,

10101101 ($l = 8$)

And consider all the *n*-grams in that generation, from 1 to $l$. For explicitness,

1-grams: [1][0][1][0][1][1][0][1]

2-grams: [10][01][10][01][01][11][10][01]

3-grams: [101][010][101][011][101]

4-grams: [1010][0101][1011][0110][1101]

…

8-grams: [10101101]

We have defined every symbol as a tree of depth 0, so considering 1-grams will not do. But, we can consider each *n*-gram for $2 \leq n \leq l$ to be a tree of depth 1 and breadth *n*. Each tree will be built with elements from the alphabet, so we have a set $T$ of trees over $\Sigma$. Since $V_T = V_N$, every indexed element from the alphabet is a leaf and a label. We have thus $n^m$ labelling possibilities, where *n* is the length of an *n*-gram and *m* is the number of distinct indexes in the alphabet. We can now build the trees of depth 1 corresponding to each of the *n*-grams, where the length of the *n*-gram is the breadth of the tree.

---

[9] A method to assess quantitative properties of strings (and which also gives some clues about the power of the procedure used to generate such strings) is given in Matlach et al. (2020), and it is based on similar principles as the ones we use here.

In concrete terms, what we mean is that the tree $T_1^2$ corresponding to the symbol [1] (i.e., the minimal tree of depth 2 and breadth 1 that contains a [1] and satisfies the NACs) can be constructed as either (20a) or (20b):

20)  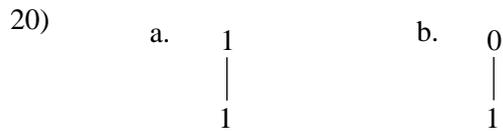
    a.   1        b.   0
          |               |
          1              1

What we are saying is that both (20a) and (20b) are well-formed trees (another way to get to the situation described in (13)). In NAC terms, what we have established is that a node labelled 1 in a tree T can dominate another node labelled 1, and that a node labelled 0 in T can dominate a node labelled 1. Bear in mind the Lonely Beta condition: 0 cannot *dominate* 0 (or, equivalently, 0 cannot precede 0 in a walk through T). Therefore, the tree

    c.   0
           |
          0

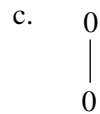

is excluded (again). This is important, since if (20c) is excluded as an elementary tree, then every derived tree in which (c) appears will also be excluded (because a well-formed tree requires the satisfaction of *all* NAC in the form of logical conjunction). Therefore, when we get to building the set of trees $T_2^2$, we consider the 2-grams [01], [10], and [11]. And here we are back at the situation specified in (14), above: a node labelled 1 can dominate a node labelled 0 or a node labelled 1; if a grammar is interpreted -as suggested by McCawley- as the *conjunction* of NACs (as opposed to their *disjunction*, which is Pullum's interpretation of what actually applies to an expression), then this means that a node labelled 1 dominates a node labelled 1 *and* a node labelled 0 (the two trees in (14), which correspond to the Fib and bif grammars, respectively).

In principle, we could define a tree of depth *n* for each *n*-gram, but that does not seem to be necessary if we allow for string concatenation / tree-composition: longer *n*-grams are simply the result of concatenating shorter *n*-grams (and therefore, of composing smaller elementary or derived trees).

We have basically done exactly the same as in **Section 2** (that is, we accomplished the same in describing the same sets of trees and providing a model for the same set of expressions), but varying the procedure a bit. The important point is that neither way of looking at things (**Section 2**'s or **Section 3**'s) requires *a priori* structural templates or makes reference to anything that is not *in the string itself*. Thus, we have defined a way to transition from superficial regularities in a string generated by the Fib grammar to a description of local allowed trees (which characterise a Fib derivation). The observations made in this paper constitute a contribution not only to the study of Lindenmayer systems as formal objects, but also to the formal foundations of experimental uses of such systems.

5. <u>Some conclusions and further issues:</u>

The goal of this paper was to provide a way to build a set of trees starting from conditions over expressions. In this sense, it is essential to note that we are not *generating* anything. There is no recursive enumeration of strings or production of strings at all; what we have done here is *model theoretic syntax* (in the sense of Pullum & Scholz, 2001). We provided arguments there is a mapping between a set of *n*-grams and a set of elementary trees. This mapping is done through constraints on possible bi-grams and tri-grams; all we have is a set of restrictions on strings, not a procedure to proof-theoretically get from strings to strings. To quote Rogers,

> *This approach abandons the notion of grammar as a mechanism and, instead, defines a language as a class of more or less ordinary mathematical structures via a linguistic theory expressed in a more or less ordinary logical language* [as opposed to rewrite mechanisms or stepwise recursive combinatorics] (1997: 722)

Both Fib and bif satisfy the First and Second Law. And the optionality between [0 1] and [1 0] in the rules arises as two equally legitimate ways to compose trees of depth 2, breadth 1, and root [1]. We have said nothing about *constituency*, which is where the differences arise (Krivochen et al., 2018). The primitives here are *n*-grams and constraints over expressions; the primitives in procedurally-based syntax are categories and rules of combination.

From the perspective of the classification of L-systems, we may note that having an L-system with a Lonely Beta is another way to say 'asymmetric L-system', as far as we can see. However, that means only that Fib and bif satisfy the same model (i.e., the same set of constraints), not that they are equivalent. Specifically, we have made no mention of *constituents* in this work, and it is constituency that defines the non-equivalence of Fib and bif; superficially, they are *almost* identical (which is precisely why bif is a good foil grammar in AGL experiments where the target grammar is Fib). If the Lonely Beta condition is dropped, or, equivalently, if we admit the 2-gram 00 while assuming the same alphabet and the Second Law, we would have 'an' XOR grammar (a symmetric grammar with the same distribution of 0 and 1 and the same growth pattern as 'the' XOR) in our hands:

21) $0 \rightarrow 1\ 0$
 $1 \rightarrow 0\ 1$

As a conclusion, we are simply pointing out that there is a way to build elementary and derived trees starting from *n*-grams *if* there are well-defined restrictions on possible *n*-grams at the local level (2- or 3-grams, we have not tried more complex grammars where larger *n*-grams would need to be considered). That is: looking at *n*-grams and looking at trees are *not* mutually exclusive things; this is an important conclusion for the analysis of L-systems. It is necessary to note that the local structures we have built (*elementary* and *derived trees*) do not correspond necessarily with *constituents* of the Fib grammar; once again, it is paramount to note that what we have done is construct a (very simple, possibly *too* simple) model that expressions of the language must satisfy.